\documentclass[runningheads]{llncs}
\usepackage{hyperref}
\usepackage{subcaption}
\usepackage{tabularx}
\usepackage{listings}
\usepackage{verbatim}
\usepackage[table]{xcolor}
\usepackage{setspace}
\usepackage{enumitem}
\usepackage{float}
\usepackage[]{caption}
\usepackage{orcidlink}
\usepackage{pdfpages}
\usepackage{enumitem}
\usepackage{graphicx}
\usepackage{booktabs}
\usepackage{amssymb}
\usepackage{rotating}
\usepackage{bbm}
\usepackage{array}
\begin{document}

\title{Re-Thinking Process Mining in the AI-Based Agents Era}
\titlerunning{Re-Thinking Process Mining in the AI-Based Agents Era}

\author{
Alessandro Berti\inst{1,2}\orcidlink{0000-0002-3279-4795},
Mayssa Maatallah\inst{3}\orcidlink{0009-0001-9966-0054},
Urszula Jessen\inst{4,5}\orcidlink{0000-0002-7282-8451},
Michal Sroka\inst{6}\orcidlink{https://orcid.org/0000-0002-7505-2521},
Sonia Ayachi Ghannouchi\inst{3}\orcidlink{0000-0001-9583-9797}
}
\authorrunning{A. Berti et al.}
\institute{Process and Data Science Chair, RWTH Aachen University, Aachen, Germany \and
Fraunhofer FIT, Sankt Augustin, Germany \and
Higher Institute of Management of Sousse, University of Sousse, Sousse, Tunisia \and
Process Insights, ECE Group Services, Hamburg, Germany \and
Eindhoven University of Technology, The Netherlands \and
Microsoft
\\
\email{a.berti@pads.rwth-aachen.de; \\
\{mayssamaatallah, ghannouchisonia.ayachi\}@isgs.u-sousse.tn; \\
u.a.jessen@tue.nl;
misroka@microsoft.com;}}
\maketitle

\begin{abstract}
Large Language Models (LLMs) have emerged as powerful conversational interfaces, and their application in process mining (PM) tasks has shown promising results. However, state-of-the-art LLMs struggle with complex scenarios that demand advanced reasoning capabilities. In the literature, two primary approaches have been proposed for implementing PM using LLMs: providing textual insights based on a textual abstraction of the process mining artifact, and generating code executable on the original artifact. This paper proposes utilizing the AI-Based Agents Workflow (AgWf) paradigm to enhance the effectiveness of PM on LLMs. This approach allows for: i) the decomposition of complex tasks into simpler workflows, and ii) the integration of deterministic tools with the domain knowledge of LLMs. We examine various implementations of AgWf and the types of AI-based tasks involved. Additionally, we discuss the CrewAI implementation framework and present examples related to process mining.
\end{abstract}

\renewcommand{\sectionautorefname}{Section}
\renewcommand{\subsectionautorefname}{Section}
\renewcommand{\subsubsectionautorefname}{Section}
\def\univs{U}
\newcommand{\univ}[1]{\univs_{\mathit{#1}}}
\newcommand{\class}[1]{\mathbb{C}_{\mathit{#1}}}
\newcommand{\pim}[1]{\pi_{\mathit{#1}}}

\let\olddefinition\definition
\renewcommand{\definition}{\small\olddefinition}


\section{Introduction}
\label{sec:introduction}

Process Mining (PM) is a branch of data science aiming to infer process-related insights starting from the event data recorded by the information systems supporting the
execution of the processes. Several types of techniques have been proposed within process mining, including process discovery (the automated discovery of a process model),
conformance checking (comparing the behavior of an event log against the process model), and predictive analytics (next activity/remaining time in a case).

Large Language Models (LLMs) have emerged as powerful PM assistants \cite{DBLP:conf/bpm/Berti0A24},
being able to: 1) effectively respond to inquiries over a textual abstraction of a PM artifact (for example, identifying semantic anomalies or root causes); 2) produce code (like Python or SQL) that can be executed over a PM artifact.

However, the implementation paradigms 1) and 2) fail in more complex scenarios.
For instance, composite tasks, which could be resolved by human analysts in different steps (for example, estimating the level of unfairness in an event log could be divided into i) identification of the protected group; ii) comparison between protected and non-protected group), are difficult for an LLM which could fail to decompose and execute each of them correctly \cite{DBLP:conf/iclr/ZhouSHWS0SCBLC23}.
Moreover, some tasks may require the production of code (to compute reliable statistics over the entire event log) but also the semantic capabilities of LLMs for the interpretation of the results obtained executing the code
(for instance, a LLM might generate some declarative process model. Then, code is generated to apply conformance checking between the log and the process model. In the last step, the semantic understanding of the LLM may be required to interpret the results).

In this paper, we propose the application of the \emph{AI-Based Agents Workflow (AgWf)} paradigm \cite{DBLP:journals/corr/abs-2310-06500} in the PM context.
AgWf(s) combine deterministic functions (called tools) and non-deterministic functions (AI-based tasks) to take the best of the two worlds:
the rich set of process mining techniques has already been developed, as well as the semantic ability of LLMs.
Moreover, AgWf(s) are deeply based on the \emph{divide-et-impera} principle, in which difficult tasks are decomposed into manageable (by the LLM) units, with the overarching goal of increasing the quality of the overall result.

\begin{figure*}[!t]
\centering
\includegraphics[width=\textwidth]{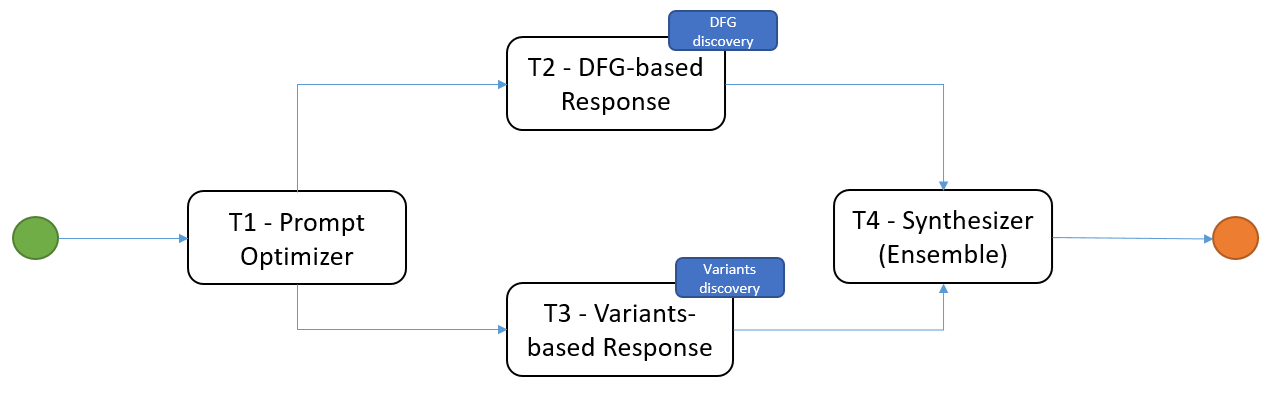}
\caption{Example AgWf, reported from Fig. \ref{fig:agwf1}, with four tasks aimed to combine the DFG and variants abstraction of an event log to respond to the inquiry of the user, which is preliminarily optimized by another task.}
\label{fig:agwf1Intro}
\vspace{-6mm}
\end{figure*}

In the following, we analyze some PM applications benefitting from an AgWf-based implementation. Moreover, we introduce some types of AI-based tasks
(routers, ensembles, evaluators, improvers) that are helpful to implement PM pipelines. We also present the CrewAI implementation framework which is helpful
to implement AgWf(s), along with two examples tailored to the PM context. In Fig. \ref{fig:agwf1Intro}, we see an example of multi-task AgWf.

The rest of the paper is organized as follows.
In Section \ref{sec:relatedWork}, we discuss the related work on PM-on-LLMs and scientific workflows in PM.
In Section \ref{sec:methodology}, we present the definition of AgWf along with a running example, possible implementations, and types of tasks.
In Section \ref{sec:implementationFramework}, we present the CrewAI framework for implementing AgWf(s).
In Section \ref{sec:nextSteps}, we discuss the future research directions in implementing AI-based agents.
Eventually, Section \ref{sec:conclusion} concludes the paper.

\section{Related Work}
\label{sec:relatedWork}

\noindent
\textbf{Process Mining on LLMs}:
In \cite{DBLP:journals/simpa/BertiZS23}, textual abstractions of PM artifacts are provided to the LLMs in order to respond to inquiries.
On the other hand, the translation of the inquiry to SQL statements executable against the original data source is proposed in \cite{DBLP:journals/corr/abs-2307-09909}, along with error-correction mechanisms.
In \cite{DBLP:conf/bpm/Berti0A24}, some types of PM that could be implemented on LLMs (semantic anomaly detection, root cause analysis, visual recognition, fairness assessments)
are discussed. Moreover, different implementation paradigms (\emph{direct provision of insights}, \emph{code generation}, and \emph{hypotheses generation}) are proposed.
In \cite{DBLP:journals/corr/abs-2407-13244}, a comprehensive benchmark for PM tasks on LLMs is proposed.

\noindent
\textbf{Connection to Traditional Process Mining Workflows}:
Scientific workflows, for example based on RapidMiner \cite{DBLP:conf/bpm/MansAV14}, Knime \cite{DBLP:conf/icpm/KouraniZLEHL22}, or SLURM \cite{DBLP:journals/corr/abs-2307-02833}, have been used in PM to increase reproducibility
and standardize large-scale experiments. However, the overarching goal of AgWf(s) is not to ensure reproducibility or standardization but to ensure the feasibility of the overall pipeline
by adopting a \emph{divide-et-impera} approach and \emph{using the right tool for every task}. Moreover, AgWf(s) are non-deterministic by nature, while for most PM workflows, the same output would be obtained starting from the same inputs (determinism).

\section{Methodology}
\label{sec:methodology}

\subsection{AI-Based Agents Workflows}
\label{subsec:aiBasedAgWf}

In Def. \ref{def:agwfDefinition}, we propose the definition of \emph{AI-Based Agents Workflow (AgWf)}. The definition contains
deterministic tools ($F$) transforming strings to other strings, and AI-based (non-deterministic) tasks ($T$).
We indicate with the symbol $\rightsquigarrow$ a non-deterministic function, providing possibly different outputs for the same input.
Each task in $T$ may be associated with a set of tools (via the ``$\textrm{tools}$'' function).
The selection of the tool for the purpose is also a non-deterministic (AI-based) function. We indicate with $\univ{\Sigma}$ the universe of all strings.

\begin{definition}[AI-Based Agents Workflow (AgWf)]\label{def:agwfDefinition} \\
An AI-Based Agents Workflow (\emph{AgWf}) is a tuple $(F, T, \textrm{tools}, \textrm{selector}, \textrm{prec}, t_1, t_f)$ for which:
\begin{itemize}
\item $F \subseteq (\univ{\Sigma} \not\rightarrow \univ{\Sigma})$ is a set of tools.
\item $T \subseteq (\univ{\Sigma} \rightsquigarrow \univ{\Sigma})$ is a set of (AI-based) tasks.
\item $\textrm{tools} : T \rightarrow \mathcal{P}(F)$ associates a set of tools to a task.
\item $\textrm{selector} : \univ{\Sigma} \times \mathcal{P}(F) \rightsquigarrow F$ selects a tool (for the given inquiry) among the available ones.
\item $\textrm{prec} : T \rightarrow \mathcal{P}(T)$ associates a task with a set of preceding tasks.
\item $t_1 \in T$ is the initial task of the workflow.
\item $t_f \in T$ is the final task of the workflow.
\end{itemize}
\end{definition}

In the definition, we never explicitly mention the term \emph{agent}. We assume that the (AI-based) \emph{agent} is the underlying executor of the (AI-based) tool.
In particular, the (AI-based) agent is involved in the execution of two different non-deterministic activities:
\begin{itemize}
\item The selection of the tool, among the available ones for the task, to be used for the purpose (while the selected tool itself is deterministic).
\item The execution of the task itself, which leads to the final response of the task.
\end{itemize}
Different (AI-based) tasks can be associated with different (AI-based) agents, accounting for the different types of skills of the agents.
For example, simpler tasks could be executed with simpler agents (decreasing the costs and the execution times), while more complex tasks require competent agents.

We also note that in Def. \ref{def:agwfDefinition}, the tasks/tool selections are non-deterministic, but the definition of the workflow is static. In Section \ref{sec:nextSteps}, we discuss the next natural step in the definition of AI-based agents: the automatic orchestration of the workflows (the tasks and their order are decided by an orchestrator).

In Def. \ref{def:agwfExecution}, we define the \emph{execution} of an AI-based workflow.
First, a sequence of tasks respecting the provided order is extracted from the workflow.
Then, each task is executed, leading to an output that is appended (via the $\oplus$ function) to the original inquiry.
While the definition in Def. \ref{def:agwfExecution} could be modified to account for concurrent executions of the tasks,
the currently available implementation framework (for AgWf) works stably within sequential executions.

\begin{definition}[AgWf Execution - Sequential]\label{def:agwfExecution} \\
Let $\textrm{AgWf} = (F, T, \textrm{tools}, \textrm{selector}, \textrm{prec}, t_1, t_f)$ be an AI-based agents workflow.
We define as execution a tuple $\textrm{ExAgWf} = (\textrm{AgWf}, S_T, S_{\Sigma})$ such that:
\begin{itemize}
\item $S_T = \langle t_1, \ldots, t_f \rangle$, with $t_1, \ldots, t_f \in T$, is a sequence of tasks respecting $\textrm{prec}$.
\item $S_{\Sigma} = \langle \sigma_0, \sigma_1, \ldots, \sigma_f \rangle$ is a sequence of strings, with $\sigma_0$ being the initial state of the workflow (i.e., the initial inquiry of the user).
\item For each $i \in \{ 1, \ldots, f \}$:
\begin{itemize}
\item If $\textrm{tools}(t_i) = \emptyset$, $\sigma_i = \sigma_{i-1} \oplus t_i(\sigma_{i-1})$.
\item If $\textrm{tools}(t_i) \neq \emptyset$, $\sigma_i = \sigma_{i-1} \oplus t_i(\sigma_{i-1} \oplus \textrm{selector}(\sigma_{i-1}, \textrm{tools}(t_i))(\sigma_{i-1}))$.
\end{itemize}
\end{itemize}
\end{definition}

In Def \ref{def:agwfExecution}, we separate between tasks without associated tools and tasks with associated tools. For the first type, the state of the workflow is defined as the concatenation of the previous state and of the result of the execution of the (AI-based) task. For the second type, we execute:
\begin{enumerate}
\item The selection of the tool, among the available ones.
\item The (deterministic) tool applied on the previous state.
\item The (AI-based) task is executed on the previous state and on the result of the application of the selected tool.
\end{enumerate}
While the output of the deterministic tool is not persisted in Def. \ref{def:agwfExecution}, it is actively used in determining the final answer of the (AI-based) task.

\begin{figure*}[!b]
\vspace{-6mm}
\centering
\includegraphics[width=\textwidth]{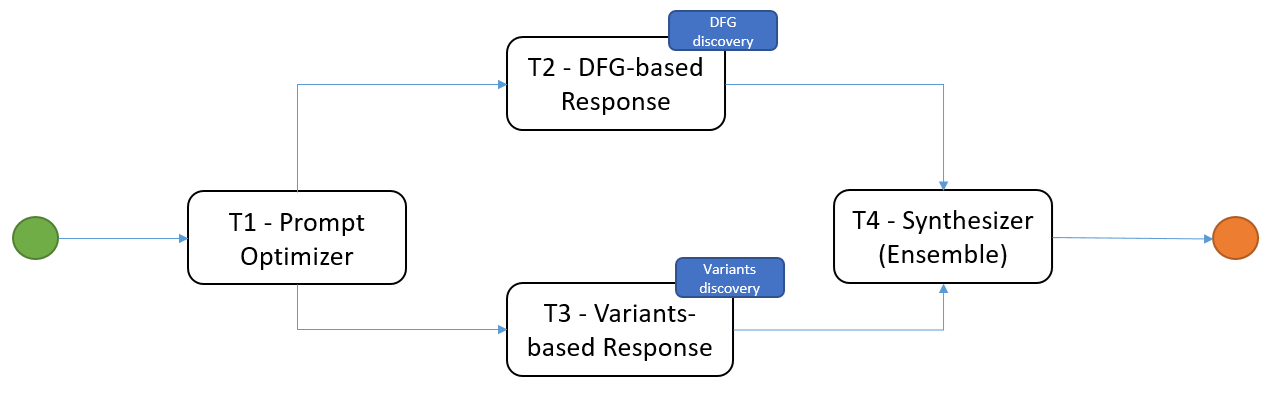}
\caption{Example AgWf in which the first task (\textbf{T1}) is optimizing the received inquiry, then two different tasks (\textbf{T2} and \textbf{T3}) are executed using two different textual abstractions (directly-follows graph and process variants) to retrieve an answer, and eventually the responses are synthesized by an ensemble (\textbf{T4}).}
\label{fig:agwf1}
\end{figure*}

\subsection{AgWf Running Example}
\label{subsec:agWfRuEx}

In Fig. \ref{fig:agwf1}, we see an example of AI-based agents workflow, aiming to exploit two different abstractions (directly-follows graph and process variants) to respond to the inquiry of the user.
The individual results are eventually merged by an ensemble, extracting the best of the single answers. The strategy, requiring the execution of four different prompts, potentially leads to better results
since different abstractions represent different views on a given PM artifact.
For example, in \cite{DBLP:conf/bpm/Berti0A23} it is discussed that the variants abstraction could be better suited for root cause analysis, while for semantic anomaly detection tasks, the knowledge of the directly-follows graph may
be sufficient. However, in some instances, the opposite choice might be more effective. For instance, in a process in which the performance problems lay in a single transition between two activities,
the DFG abstraction might be more effective in showing the root cause of the performance issue.
On the other hand, in a P2P process, if an invoice is paid twice non-consecutively, that would be hidden in the directly-follows graph abstraction but would be visible in the variants abstraction.

Considering always Fig. \ref{fig:agwf1}, we see clear start/end tasks (\textbf{T1} and \textbf{T4}). Two different sequences of tasks are allowed by the workflow (since \textbf{T2} and \textbf{T3} are interleaved):
$\langle \textbf{T1}, \textbf{T2}, \textbf{T3}, \textbf{T4} \rangle$ and $\langle \textbf{T1}, \textbf{T3}, \textbf{T2}, \textbf{T4} \rangle$.
The execution of each task appends the result to the input string. For instance, an initial inquiry of the user, \emph{Tell me the violations in the process contained in the event log at} \textbf{/home/erik/p2p.xes}, can be optimized by \textbf{T1}
appending \emph{Could you analyze the behavior in the process, providing a list of anomalous behavior?}. Then \textbf{T2} would append its analysis based on the DFG, for instance, \emph{``Create Purchase Requisition''
should never transition to ``Create Purchase Order'' without approval}. \textbf{T3}, on the other hand, would exploit the behavior evidenced in the variants, appending \emph{``You should never pay twice the same invoice''}.
Eventually, \textbf{T4} would provide a composition of the two provided insights, appending \emph{``In conclusion, the main problems are the lack of standardization in the management of purchase requisitions and multiple payments for the same invoice.''}.

Always in Fig. \ref{fig:agwf1}, we see that \textbf{T2} and \textbf{T3} are annotated with some tools, i.e., \emph{DFG discovery} and \emph{Variants discovery}.
For instance, these can be implemented using the \emph{pm4py} process mining library \cite{DBLP:journals/simpa/BertiZS23}, opening the XES referenced in the input string and applying a discovery operation.
The results of the tools are appended to the prompt before the (AI-based) task is executed. Tools help to avoid ``re-inventing the wheel'', releasing the task from the duty to compute
the DFG/process variants\footnote{Assuming that the provision of the entire event log in the prompt is feasible, which in many cases is not, due to the limited length of strings accepted
by the AI-based tasks.}.

\begin{figure*}[!b]
\vspace{-2mm}
\centering
\includegraphics[height=0.07\textheight]{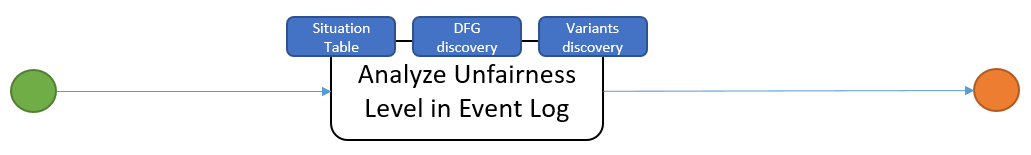}
\caption{Example AgWf for bias detection in process mining (single task - multiple tools).}
\label{fig:agwf2_1}
\end{figure*}

\begin{figure*}[!b]
\vspace{-2mm}
\centering
\includegraphics[height=0.07\textheight]{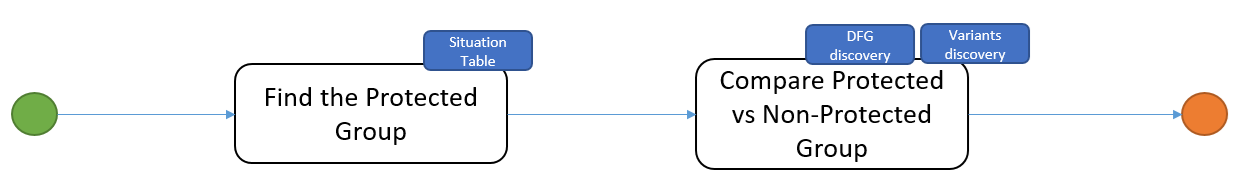}
\caption{Example AgWf for bias detection in process mining (multiple tasks - multiple tools).}
\label{fig:agwf2_2}
\end{figure*}

\begin{figure*}[!b]
\vspace{-5mm}
\centering
\includegraphics[width=\textwidth]{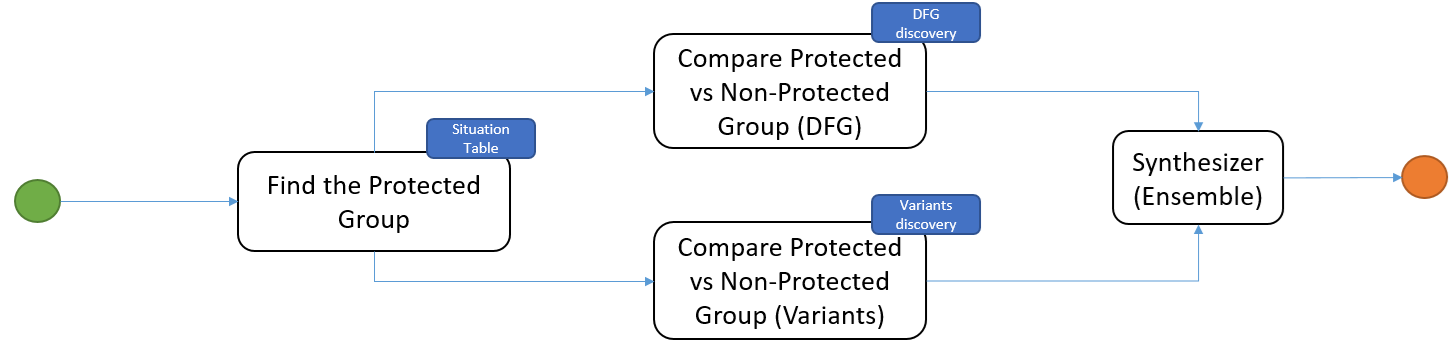}
\caption{Example AgWf for bias detection in process mining (multiple tasks - single tool per task).}
\label{fig:agwf2_3}
\end{figure*}

\subsection{Possible Implementations}
\label{subsec:possibleImplementations}

Several possible implementations of AgWf(s) are possible for the same tasks 
In Fig. \ref{fig:agwf2_1}, Fig. \ref{fig:agwf2_2}, and Fig. \ref{fig:agwf2_3}, we see different implementations of AgWf for the same problem (bias detection).
However, arguably, there is a clear rank of effectiveness in the implementations.

The least effective implementation is represented in Fig. \ref{fig:agwf2_1}. There is a single task, whose final output should be an estimation of the unfairness level
in the considered event log. The task would be resolved by a human analyst as follows \cite{DBLP:conf/bpm/PohlBQA23}:
\begin{itemize}
\item The event log is divided between cases belonging to the protected group and cases belonging to the non-protected group.
\item The behavior is compared between the protected and non-protected groups.
\end{itemize}
However, AI-based tasks as in Definition \ref{def:agwfDefinition} are allowed to use a single tool and produce a one-shot response to the provided inquiry.
Therefore, they would not be able to perform successfully the pipeline. In the best case, they would be able to infer some fairness-related insights from the process variants
(e.g., when activities potentially related to discriminations are contained in the process variants).

The workflow proposed in Fig. \ref{fig:agwf2_2} is decomposed into two tasks, identification of the protected group and comparison between the protected versus the non-protected group.
The pipeline is sound. However, the second task needs to make a choice between several tools, which could lead to incomplete fairness insights. 

The workflow proposed in Fig. \ref{fig:agwf2_3} is decomposed into four tasks. The second and third tasks compare the protected and non-protected groups in two different ways,
based on the DFG and process variants abstractions. The results are then collated by the ensemble to form the final report on the unfairness contained in the event log.

Since current AI-based agents are still limited in their performance and scope of action, the decomposition of the final goal (shown in Fig. \ref{fig:agwf2_1}) into different sub-tasks,
each limited in scope, leads more straightforwardly to the desired output.

\subsection{Types of Tasks}

\begin{figure*}[!b]
\vspace{-5mm}
\includegraphics[width=\textwidth]{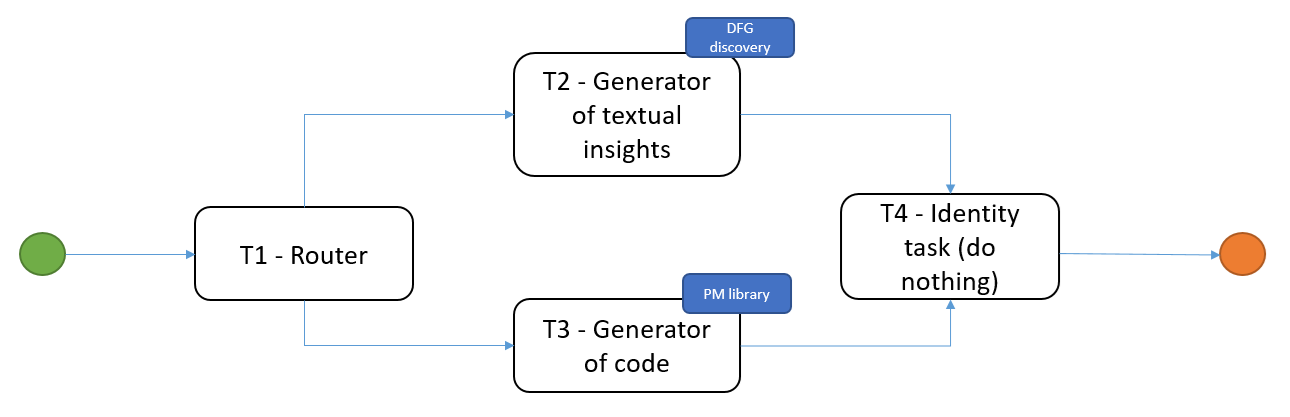}
\caption{An AgWf in which an inquiry is redirected either to \textbf{T2} (LLM-based textual insights, i.e., checking the semantics) or \textbf{T3} (production of code that is executed
against the log in an existing PM library).}
\label{fig:routers}
\end{figure*}

In Fig. \ref{fig:agwf1} and Fig. \ref{fig:agwf2_3}, we show the role of tasks in AgWf workflows. In this section, we aim to discuss different types of tasks and their utility, especially
focusing on the PM context.
\begin{itemize}
\item \emph{Prompt Optimizers}: tasks accepting the original inquiry of the user and transforming it to a language tailored to the capabilities of the AI agents.
They usually are not associated with any tool, as their role is to optimize the cleanliness and effectiveness of the inquiry.
\item \emph{Ensembles}: tasks accepting a prompt containing a collation of insights (collected from different tasks offering a different perspective) and returning a coherent
text containing the main results of the analysis. For example, the ensemble could summarize analyses over different dimensions (control-flow, temporal, data, resource) into
a unified report on the process.
\item \emph{Routers}: tasks accepting a prompt and deciding which one of the depending nodes should be executed. While explicit routing is not allowed within the context
of Def. \ref{def:agwfDefinition}, the following tasks could be instructed to consider the output of the routing node and possibly skip the production of further output.
For example, in Fig. \ref{fig:routers}, we see a typical routing decision, i.e., choosing if the problem should be resolved directly by the LLM as it is a semantical task
and/or it does not require extensive access to the attributes of the event log, or generating some code executable against the event log using a PM library such as
\emph{pm4py}.
\item \emph{Evaluators}: tasks evaluating the output of a previous task and assessing the quality, for instance, assigning a score between 1.0 and 10.0. This might help to understand the effectiveness of the task's execution. While the definition of \emph{AgWf} does not allow for loops, in case of outputs with low quality is it possible to implement a ``wrap back'' mechanism
in which the execution is taken back to a previous state and repeated.
\item \emph{Output Improvers}: trying to enhance the quality of the output of the previous tasks. For instance, the insights could be refined (``second opinion'') or, in the case of
code generation, the quality or security of the code can be improved.
\end{itemize}

\section{Implementation Framework}
\label{sec:implementationFramework}

In this section, we present the \emph{CrewAI} Python framework \url{https://github.com/crewAIInc/crewAI} for the implementation of AgWf(s) on top of Large Language Models. It is based on the following concepts:
\begin{itemize}
\item \emph{AI-based agents} are defined as LLMs plus system prompts. The system prompt tailors the behavior of a given LLM to a given role (\emph{role prompting} \cite{DBLP:journals/corr/abs-2403-02756}).
\item \emph{AI-based tasks} are defined based on a textual instruction. They are associated with an AI-based agent.
\item \emph{Tools} are defined as Python units (classes/functions). A task can be connected to some tools. The selection operates on the documentation string of the different tools, including the input arguments and the type of the output.
\item In the traditional implementation, a sequential order of execution for the tasks is defined. More recently, a concept of concurrent execution (\emph{hierarchical processes}) has been tried, but further work is needed.
\end{itemize}
An important criteria for the selection of the LLM is its ability as selector for the most suitable tool. Notably, LLMs such as \emph{Llama-3.1}\footnote{\url{https://ai.meta.com/blog/meta-llama-3-1/}},
\emph{Qwen 2.0}\footnote{\tiny \url{https://medium.com/@smalltong02/the-comprehensive-evaluation-of-the-agency-capability-of-the-qwen-2-model-cb7eb675c091}},
\emph{Mistral Large 2}\footnote{\url{https://techcommunity.microsoft.com/t5/ai-machine-learning-blog/ai-innovation-continues-introducing-mistral-large-2-and-mistral/ba-p/4200181}}
or \emph{GPT-4O/GPT-4O-Mini}\footnote{\tiny \url{https://cobusgreyling.medium.com/langchain-based-plan-execute-ai-agent-with-gpt-4o-mini-243ee57c6a5a}} offer excellent support to implement AgWf.
Also, since a workflow contains potentially many different tasks, the speed of the model is important.
For instance, \emph{Llama 3.1 70B} and \emph{GPT-4O-Mini} could be a preferred choice over their bigger siblings \emph{Llama 3.1 405B} and \emph{GPT-4O} due to their satisfactory performance
at a lower computational price.

CrewAI supports also additional concepts in comparison to Def. \ref{def:agwfDefinition}:
\begin{itemize}
\item The \emph{entity memory} is a dictionary persisting variables produced/accessed during the execution of the workflow. For instance, from an initial log \textbf{log}, we could create two sub-logs (for example, \textbf{training} and \textbf{test}). The \textbf{training} log could be then accessed to generate some hypotheses that are then tested on the \textbf{test} log.
\item Python functions (\emph{callbacks}) could be called at the end of the execution of some tasks (for example, to persist the result, or check the formal correctness).
\end{itemize}

In the following, we will propose two examples of AI-based workflows.
We implemented in CrewAI the fairness workflow shown in Fig. \ref{fig:agwf2_2}. A Jupyter notebook is available at the address \url{https://github.com/fit-alessandro-berti/agents-trial/blob/main/02\_fairness\_assessment.ipynb}.
The selected LLM for the task is \emph{Qwen 2.0 72B}.
In particular, the first task (identification of the protected group) generates some code (SQL statement) executed against the event log in order to split the behavior between ``protected'' and ``non-protected'' cases.
Using the advanced features provided by CrewAI, both event logs are stored in the entity memory for usage in the following task.
The following task (comparison between protected and non-protected groups) calls the computation of the DFG on both event logs and returns a textual list of insights.
We see that we assign each task to a different \emph{agent}. Despite both agents are being supported by the same LLM (\emph{Qwen 2.0 72B}), the system prompt defines a different purpose
for each LLM. The tasks are defined, following the CrewAI framework implementation, with a \emph{description} and an \emph{expected output}.

\begin{figure*}[!t]
\centering
\includegraphics[width=\textwidth]{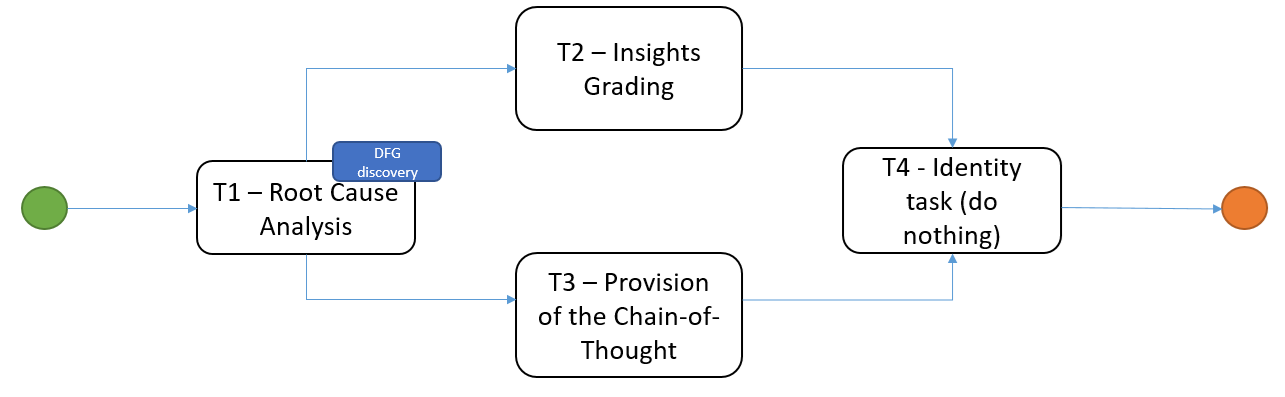}
\caption{Workflow for root cause analysis, including the insights grading and the provision of the chain-of-thought as evaluation steps.}
\label{fig:woFloRca}
\vspace{-5mm}
\end{figure*}

We also define another AgWf for root cause analysis, including two different mechanisms of evaluation, at the address \\
\url{https://github.com/fit-alessandro-berti/agents-trial/blob/main/01_root_cause_analysis_insights.ipynb}.
The workflow is shown in Fig. \ref{fig:woFloRca}. In particular, the first step \textbf{T1} performs root cause analysis starting from the DFG abstraction, producing a list of potential root causes.
Then, \textbf{T2} assigns to each insight a confidence score \cite{DBLP:journals/corr/abs-2402-12563} from 1.0 (minimum) to 10.0 (maximum). \textbf{T3} needs to provide the chain-of-thought \cite{DBLP:conf/nips/Wei0SBIXCLZ22} for the first of the provided insights,
so the detailed reasoning steps. The graded insights and the detailed reasoning steps (for the first insight) are then returned at the end of the workflow.
We see that we also define three different agents on the same LLM with different system prompts.
In comparison to the previously defined workflow, the tasks are easier to perform. Therefore, we propose to use the \emph{Qwen 2.0 8B} LLM, which is significantly smaller than \emph{Qwen 2.0 72B} LLM.

\section{Next Steps}
\label{sec:nextSteps}

\noindent
\textbf{Automatic Definition of AgWf(s)}:
In the previous sections, we show how tasks could be decomposed into an AgWf. However, decomposition is done by humans. Some approaches \cite{DBLP:journals/corr/abs-2402-16713,DBLP:journals/corr/abs-2405-16334} show that the same decomposition task could be performed by an ``orchestrating'' LLM. In particular, the original task is decomposed into a sequence of smaller tasks assigned to specialized agents. One of the main challenges identified in \cite{DBLP:journals/corr/abs-2402-16713} is the comprehension of the original task. In particular, it is argued that the orchestrating LLM should be instructed to request clarifications on the task.

\noindent
\textbf{Tasks Keeping the Human-in-the-Loop}: AgWf(s) could be used to automate many tasks. However, the execution of some tasks might benefit from clarifications provided by the end user \cite{abuelsaad2024agent}. For instance, the prompt optimizer depicted in Fig. \ref{fig:agwf1} could struggle to optimize a very generic inquiry (such as ``What are the problems in the process?''), and could benefit from additional clarifications provided by the user.

\noindent
\textbf{Evaluating AgWf(s)}:
In this paper, we argue that AgWf(s) are useful tools to increase the quality of the output on a specific task by decomposing it into smaller tasks executed by specialized agents.
The assessment of LLM-based outputs is challenging, with the \emph{LLMs-as-a-Judge} paradigm \cite{DBLP:journals/corr/abs-2403-02839} being a popular solution. Within AgWf(s), the overall effectiveness (quality of the final output) depends on the effectiveness of the single agents. For instance, errors in the initial routing of the inquiry can result in a significantly lower quality output even if all the other tasks are performed optimally. Therefore, we argue that the quality of the output of the single tasks should be assessed.

Also, when multiple agents are involved, the collaboration and psychological traits should be considered \cite{DBLP:journals/corr/abs-2310-02124,DBLP:journals/corr/abs-2402-08189}. Tasks could be implemented with self-awareness or also awareness of the context (overarching goal/workflow context). In \cite{DBLP:journals/corr/abs-2404-16698}, it is argued that even currently top-performing LLMs show poor cooperation behavior/negotiation skills, leading sometimes to poor outcomes.

\noindent
\textbf{Maturity of the Tool Support}: several frameworks are proposed to implement AgWf(s). In particular, we need to mention the \emph{LangGraph} framework \url{https://langchain-ai.github.io/langgraph/}, which offers comprehensive support for tasks/tools (for instance, direct connection to the search engines is provided). However, due to the ever-evolving structure of the library, prototypes depend on a specific version of the library and can stop working on a future version. Also, some user interfaces have been proposed for LangGraph\footnote{For example, \url{https://github.com/LangGraph-GUI/LangGraph-GUI}},
but they are also highly dependent on the version of the underlying library.
\emph{CrewAI} has been proposed in the paper as a ``compromise'' solution between ease-of-use and features support. However, it still lacks some advanced features offered by LangGraph and lacks a graphical interface.
The \emph{AutoGen} solution by Microsoft \url{https://microsoft.github.io/autogen/} is also a reasonable choice but lacks the completeness of LangGraph. It comes with a user-friendly graphical interface allowing for the definition of the workflows.
Overall, all the considered solutions need further work to reach high maturity.

\section{Conclusion}
\label{sec:conclusion}

In this paper, we analyzed the limitations of the currently proposed implementations for PM-on-LLMs, proposing AgWf(s) as a possible solution involving i) the decomposition of the original task in smaller units; ii) the combination between AI-based task execution and ``deterministic'' tools (for instance, using the features offered by process mining libraries).
AgWf(s) have a different goal (i.e., maximizing the quality of the output) than previously proposed scientific workflows (i.e., allowing the reproducibility of scientific experiments).
We propose different types of AI-based tasks useful for process mining applications, including prompt optimizers, ensembles, routers, evaluations, and output improvers. We use the \emph{CrewAI} framework to implement some example AgWf(s) (including root cause analysis and bias detection in process mining event logs). We also discuss future directions for research and development, including the automatic definition of the workflows, evaluation frameworks for agents, and increased maturity of the underlying frameworks/tool support.
Overall, AgWf(s) offer a powerful tool for PM-on-LLMs, requiring a \emph{divide-et-impera} mindset.

\bibliographystyle{splncs04}
\bibliography{references}

\end{document}